\documentclass[conference]{IEEEtran}
\usepackage{cite}
\usepackage{amsmath,amssymb,amsfonts}
\usepackage{algorithmic}
\usepackage{graphicx}
\usepackage{textcomp}
\usepackage{xcolor}
\usepackage{algorithm}

\usepackage{times}
\usepackage{soul}
\usepackage{url}
\usepackage[hidelinks]{hyperref}
\usepackage[utf8]{inputenc}
\usepackage[small]{caption}
\usepackage{amsmath}
\usepackage{amsthm}
\usepackage{booktabs}

\def\BibTeX{{\rm B\kern-.05em{\sc i\kern-.025em b}\kern-.08em
    T\kern-.1667em\lower.7ex\hbox{E}\kern-.125emX}}

\definecolor{Wgreen}{HTML}{01a049}

\newcommand{\warn}[1]{}

\usepackage{url}
\begin{document}

\title{TwinSegNet: A Digital Twin-Enabled Federated Learning Framework for Brain Tumor Analysis}

\author{
    \IEEEauthorblockN{Almustapha A. Wakili, Adamu Hussaini, Abubakar A. Musa,  Woosub Jung, and Wei Yu}
    \IEEEauthorblockA{
        Department of Computer Science and Information Sciences, Towson University, USA  \\
        Emails: \{awakili1,\,ahussa7,\,aahmadm1\}@students.towson.edu, \{woosubjung,\,wyu\}@towson.edu
    }
}

\maketitle

\begin{abstract}
Brain tumor segmentation is critical in diagnosis and treatment planning for the disease. Yet, current deep learning methods rely on centralized data collection, which raises privacy concerns and limits generalization across diverse institutions. In this paper, we propose \textbf{TwinSegNet}, which is a privacy-preserving federated learning framework that integrates a hybrid ViT-UNet model with personalized digital twins for accurate and real-time brain tumor segmentation. Our architecture combines convolutional encoders with Vision Transformer bottlenecks to capture local and global context. Each institution fine-tunes the global model of private data to form its digital twin. Evaluated on nine heterogeneous MRI datasets, including BraTS 2019–2021 and custom tumor collections, TwinSegNet achieves high Dice scores (up to 0.90\%) and sensitivity/specificity exceeding 90\%, demonstrating robustness across non-independent and identically distributed (IID) client distributions. Comparative results against centralized models such as TumorVisNet highlight TwinSegNet’s effectiveness in preserving privacy without sacrificing performance. Our approach enables scalable, personalized segmentation for multi-institutional clinical settings while adhering to strict data confidentiality requirements.
\end{abstract}


\section{Introduction}
Brain tumors are one of the most serious and life-threatening diseases, requiring precise detection and segmentation for effective treatment planning. Magnetic Resonance Imaging (MRI) is the preferred imaging modality for diagnosing brain tumors because of its high-resolution soft tissue contrast.  Nonetheless, manual tumor segmentation is time-consuming and inefficient. The results are subject to inter-observer variability and are heavily dependent on the radiologists' expertise. This has driven the development of automated deep learning (DL)-based segmentation models to assist in tumor detection and classification~\cite{wang2022deep}.

Despite the progress in deep learning-enabled medical image analysis, several challenges persist in MRI-based brain tumor segmentation: (i) {\em Data Privacy and Regulatory Constraints:} Medical image datasets are often distributed across multiple hospitals and research institutions. Due to strict privacy regulations (e.g., HIPAA, GDPR), sharing raw patient data for centralized AI training is not feasible. (ii) {\em Model Generalization Issues:} Most deep learning-based segmentation models are trained on limited, single-institution datasets, resulting in poor generalization across diverse patient populations. (iii) {\em Computational Complexity:} Advanced model-based segmentation models, especially those leveraging Transformer models, are computationally intensive, thereby challenging real-time applications in a clinical environment. (iv) {\em Absence of Personalized Segmentation Models:} Federated deep learning models operating in network environments in existence today do not dynamically adjust to inter-hospital differences, thus restricting their efficiency in actual clinical practice.

To overcome these challenges, digital twin (DT) and federated learning (FL) have emerged as powerful tools for privacy-preserving AI-driven healthcare supported by networking environments. DT creates patient-specific virtual models, allowing real-time disease modeling and simulation~\cite{li2023dtbvis,wentzel2024ditto}. In the medical field, FL enables collaborative AI model training among multiple parties without disclosing patient data, thereby ensuring data privacy and regulatory compliance~\cite{jooarticlesblockchain,fi17030107,sultanpure2024internet}. However, we acknowledge that vulnerabilities exist that can be exploited to infer or reverse-engineer such information, and these limitations should be considered when applying FL in practice. Recent studies have integrated DT with FL to enhance the analysis of brain tumors. For example, Sultanpure {\em et al.}~\cite{sultanpure2024internet} demonstrated how IoT-driven DT could facilitate real-time MRI-based tumor segmentation, while Joo {\em et al.}~\cite{jooarticlesblockchain} explored privacy-aware FL models for training deep learning models across multiple healthcare institutions. However, most of these approaches lack robust personalization mechanisms and are not optimized for real-time clinical deployment.

To deal with these challenges, we propose TwinSegNet, a novel Privacy-Preserving FL Framework that integrates a hybrid ViT-UNet Model for precise and efficient MRI-based brain tumor segmentation and privacy-preserving FL to train AI models across multiple hospitals while ensuring data confidentiality. Our framework also considers DT personalization, which is achieved through client-specific fine-tuning of the global model on local institutional data, enabling adaptive and real-time tumor segmentation tailored to each hospital’s patient distribution. 

The major contributions of this work are: (i) {\em Federated ViT-UNet for Brain Tumor Segmentation:} We develop a privacy-preserving hybrid ViT-UNet model trained in a FL setting operated by networks, allowing collaborative training across institutions without raw data sharing. (ii) {\em DT Personalization via Local Fine-Tuning:} We implement DT as client-specific models derived by fine-tuning the global model on each client’s private dataset after each learning round. This improves segmentation performance without requiring data centralization. (iii) {\em Privacy-Preserving FL Architecture:} We implement a secure FL protocol to facilitate multi-institutional AI training, ensuring privacy preservation and regulatory compliance. (iv) {\em Computational Efficiency Optimization for Real-Time Deployment:} We optimize ViT-UNet’s computational performance, reducing inference time and making it feasible for real-world clinical use.

The remainder of this paper is organized as follows. Section~\ref{sec:background} provides background on FL, DT, and brain tumor segmentation. Section~\ref{sec:approach} details the proposed TwinSegNet framework, including its model architecture, training strategy, and digital twin integration. Section~\ref{sec:evaluation} presents and analyzes experimental results. Finally, Section~\ref{sec:conclusion} concludes the paper.

\section{Background}
\label{sec:background}

A brain tumor is an abnormal mass of tissue within the brain or spinal canal, which can originate locally or spread from other body parts (metastatic)~\cite{aans2025brain}. Symptoms depend on tumor type, size, and location, including headaches, seizures, and speech or memory impairments. Diagnosis typically involves MRI/CT scans, neurological exams, and sometimes biopsy~\cite{weller2014eano}. Emerging AI technologies, including DT and FL, enhance precision medicine by improving diagnosis and treatment planning~\cite{awujoola2024advancing}.

FL enables decentralized training across distributed data sources, preserving privacy by sharing model parameters rather than raw data~\cite{hatcher2018survey}. In healthcare, this allows hospitals to train models while keeping patient data local and collaborative. FL mitigates the limitations of traditional centralized learning approaches that conflict with data privacy regulations such as HIPAA and GDPR~\cite{pene2023incentive}.

DT is a virtual replica of the physical system linked in real-time via sensors and actuators~\cite{hussaini2022taxonomy,hussaini2023digital}. DTs enable continuous data-driven simulations, supporting real-time monitoring, prediction, and optimization across various domains, including healthcare, manufacturing, and energy~\cite{holopainen2024digital}. In healthcare, DT can model patient conditions, forecast disease progression, and guide personalized treatment decisions~\cite{kamel2021digital,qian2024new}.

\section{Our Approach}
\label{sec:approach}

\subsection{Design Rationale}
While FL and DT have individually shown promise for privacy-aware and personalized healthcare, their isolated use fails to address the full spectrum of challenges in brain tumor segmentation. Existing FL models often lack adaptability to diverse institutional datasets, resulting in poor generalization in non-IID settings. Similarly, DT implementations rarely scale across hospitals or include deep learning-driven segmentation pipelines. The proposed FL framework operates in a distributed and federated networking environment, where multiple decentralized clients (e.g., hospital servers or edge devices) communicate over secure, low-latency network links to collaboratively train a global model. Providing effective network communication protocols and seemingly coordinated operations is vital for ensuring scalable FL model aggregation, synchronization, and privacy-preserving data exchange across heterogeneous edge nodes, which enables robust and adaptable performance in real-world, multi-institutional healthcare networks.

Given the limitations of existing DT and FL models, TwinSegNet is proposed as an end-to-end hybrid ViT-UNet framework that integrates (i) A privacy-preserving FL architecture for distributed MRI-based brain tumor segmentation, (ii) Hybrid ViT-UNet deep learning models to enhance segmentation accuracy, (iii) DT integration for real-time patient-specific tumor analysis. By leveraging DT, FL, and Vision Transformers, TwinSegNet aims to achieve (i) Real-time, privacy-aware brain tumor segmentation without data centralization, (ii) Improved tumor boundary detection via multi-modal MRI fusion models, (iii) Large-scale, cross-institutional AI training without any compromise to patient confidentiality. 

\begin{figure}[htbp]
    \centering
    \includegraphics[width=0.98\linewidth]{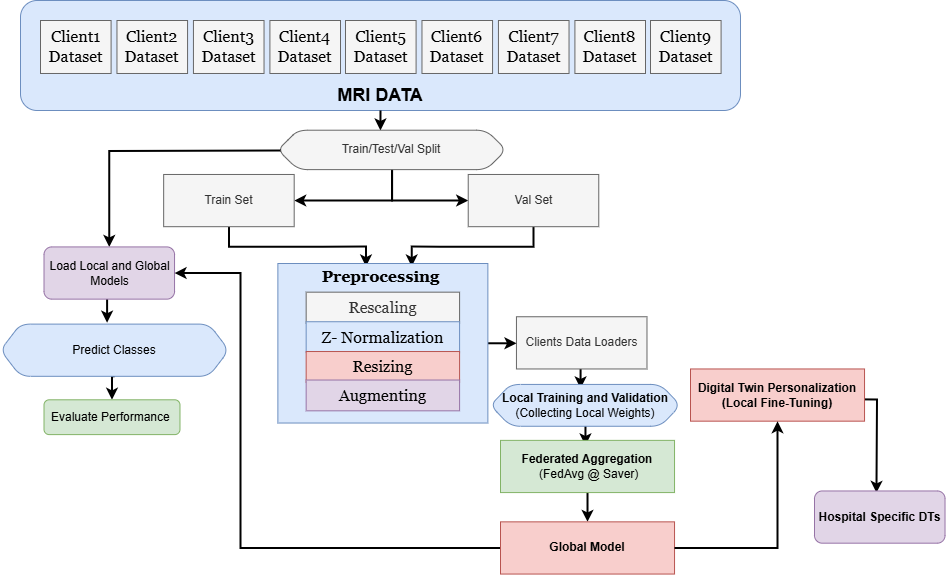}
    \caption{Workflow of TwinSegNet.}
    \label{fig:project_workflow}
\end{figure}

Fig.~\ref{fig:project_workflow} gives an overview of the proposed TwinSegNet system. The workflow consists of dataset preparation, preprocessing, model architecture, DL framework, and DT integration for personalized tumor segmentation. To be specific, the workflow begins with raw MRI datasets distributed across nine hospitals (clients). Each dataset undergoes a train-test-validation split, followed by a standardized preprocessing stage involving intensity rescaling, $Z$-normalization, spatial resizing, and data augmentation. Processed data is then passed through client-specific data loaders for local training using the TwinSegNet architecture. Each client trains the model independently on their respective data, producing local weight updates. These updates are transmitted to a central server, where federated aggregation (via FedAvg) produces a new global model.

As a setup of FL, to improve institutional adaptation, each client fine-tunes the global model based on its local validation data, resulting in a hospital-specific digital twin model. This model is subsequently used to predict tumor classes and evaluate segmentation performance at each site. The architecture ensures privacy preservation while supporting personalized and adaptive tumor analysis across diverse data distributions.

\subsection{Dataset Description}
TwinSegNet is evaluated on a composite of nine distinct brain tumor MRI collections, each assigned to a separate federated client representing an individual hospital. This design simulates a real-world decentralized healthcare setting, where institutions possess heterogeneous, non-identically distributed (non-IID) patient populations while retaining data privacy.

The datasets originate from internationally recognized medical imaging repositories and challenges: (i) \textbf{BraTS Datasets (2019–2021):} Provided by the Multimodal Brain Tumor Segmentation Challenge (BraTS) organized by MICCAI, these datasets focus on glioma segmentation with multimodal MRI inputs (T1, T1ce, T2, FLAIR). They are curated by the University of Pennsylvania and collaborating institutions to advance research in brain tumor segmentation. (ii) \textbf{2023GLI, 2023MEN, 2023MET, 2023PED, 2023SSA Collections:} Independently curated, these datasets extend beyond gliomas, covering Meningiomas, Metastatic tumors, Pediatric brain tumors, and Secondary tumor types. They enable a broader generalization of models across various tumor histologies and patient demographics.

Each hospital client uses a distinct dataset, introducing diversity in scanner protocols, imaging quality, tumor histology, and patient demographics. Table~\ref{tab:dataset_summary} summarizes the datasets assigned to each hospital along with their tumor focus and sample sizes.

\begin{table}[h]
\centering \footnotesize
\caption{Dataset-to-Client Assignment and Characteristics.}
\label{tab:dataset_summary}
\resizebox{0.5\textwidth}{!}{%
\begin{tabular}{|p{1cm}|p{2.3cm}|p{2.8cm}|p{1.8cm}|}
\hline
\textbf{Clients} & \textbf{Datasets} & \textbf{Tumor Types} & \textbf{No. of Patients} \\ \hline
Hospital1 & 2023GLI & Glioma & 1251 \\ \hline
Hospital2 & 2023MEN & Meningioma & 1000 \\ \hline
Hospital3 & 2023MET & Metastatic Tumor & 165 \\ \hline
Hospital4 & 2023PED & Pediatric Tumor & 99 \\ \hline
Hospital5 & 2023SSA & Secondary Tumor & 60 \\ \hline
Hospital6 & BraTS 2021 & Glioma (HGG and LGG) & 1251 \\ \hline
Hospital7 & BraTS 2020 & Glioma & 369 \\ \hline
Hospital8 & BraTS 2019 (HGG) & High-Grade Glioma & 259 \\ \hline
Hospital9 & BraTS 2019 (LGG) & Low-Grade Glioma & 76 \\ \hline
\end{tabular}
}
\end{table}

This federated configuration creates a naturally non-IID distribution among clients, with variability in MRI scanner protocols, patient cohorts, and tumor characteristics. Each client retains its dataset locally, training its TwinSegNet model without sharing raw imaging data, thereby preserving privacy in accordance with HIPAA and GDPR. All MRI samples have four modalities: T1, T1ce, T2, and FLAIR. Voxel-wise ground truth annotations define four segmentation classes: background, peritumoral edema, enhancing tumor core, and non-enhancing tumor core.

\subsection{Data Preprocessing}

The MRI volumes are preprocessed using the TorchIO library with the following steps: (i) {\em Rescale Intensity:} Normalize voxel intensities to $[0,1]$. (ii) {\em Z-Normalization:} Standardize intensities to zero mean and unit variance. (iii) {\em Resizing:} Resample to a uniform $128 \times 128 \times 128$ volume size. (iv) {\em Data Augmentation:} Apply random elastic deformations, bias fields, noise injection, affine transformations, and flips during training to enhance model generalization.

\subsection{Learning Model Architecture}

\begin{figure}[htbp]
    \centering
    \includegraphics[width=0.98\linewidth]{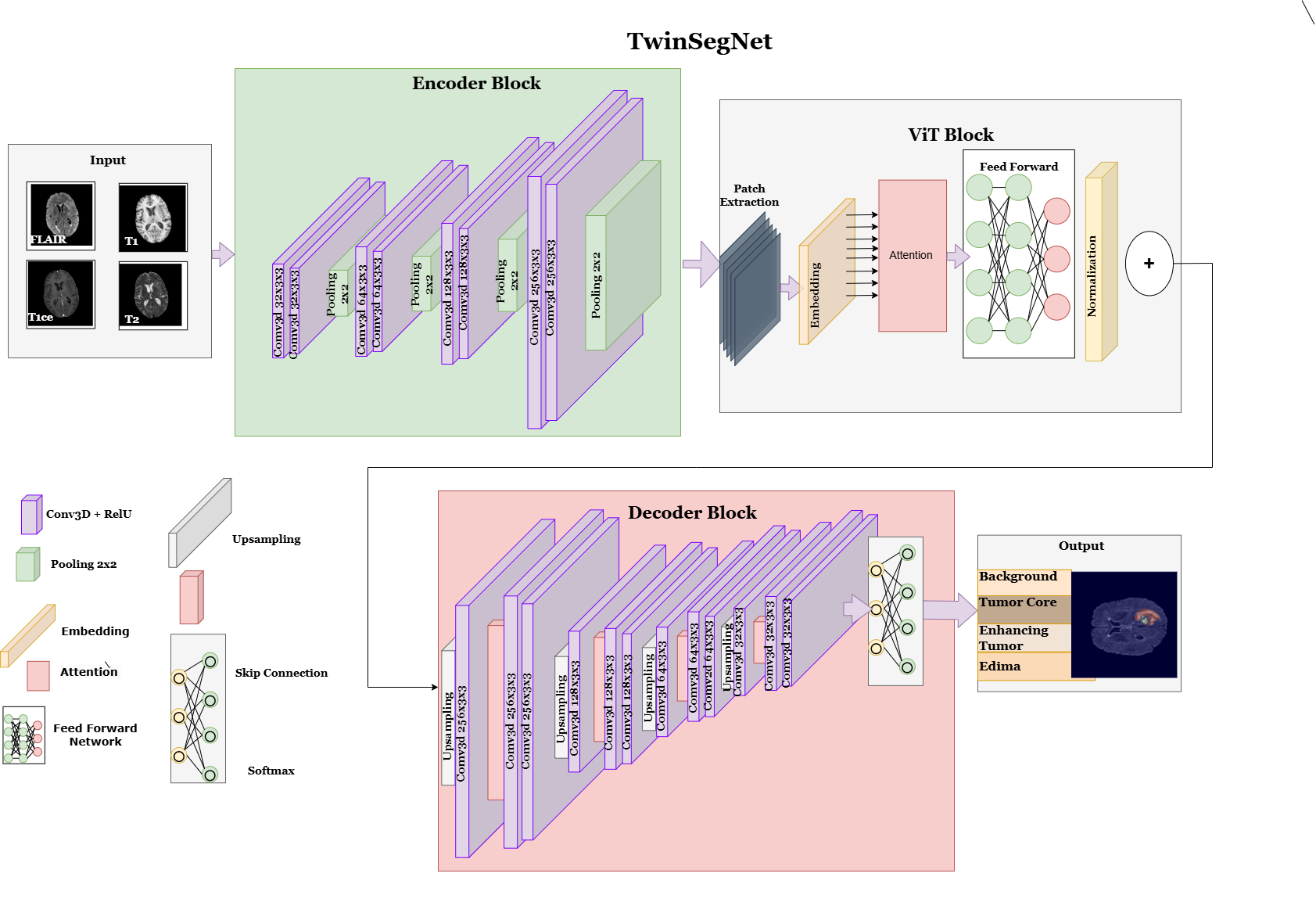}
    \caption{Hybrid Learning Architecture of TwinSegNet.}
    \label{fig:twinsegnet_architecture}
\end{figure}

Concerning the learning model, TwinSegNet considers a hybrid architecture that combines Convolutional Neural Networks (CNNs) for local feature extraction and Vision Transformers (ViTs) bottleneck for global context modeling. It is organized in a UNet~\cite{ronneberger2015u} inspired encoder-decoder structure. Fig.~\ref{fig:twinsegnet_architecture} shows the architecture of TwinSegNet. As seen in the figure, it fuses a convolutional encoder-decoder structure with a Vision Transformer (ViT) bottleneck. Four MRI modalities (FLAIR, T1, T1ce, T2) are input. Hierarchical features are encoded via Conv3D blocks, which are input to a ViT block to model global attention. The decoder upsamples and employs skip connections to reconstruct segmentation masks. Voxel-wise labels for background, tumor core, enhancing tumor, and edema are output.

In the following, we briefly outline the key components of the learning architecture of TwinSegNet: (i) {\em Encoder:} It consists of stacked 3D convolutional blocks, comprising two convolutional layers, batch normalization, and ReLU activation. Spatial resolution is decreased step by step through max-pooling layers, thereby increasing the depth of feature representation. (ii) {\em Transformer Bottleneck:} A lightweight 3D Patch Embedding ViT module operates at the network's bottleneck stage. Feature maps are divided into patches, embedded, and processed using multi-head self-attention to catch long-range dependencies. Positional embeddings are included to retain spatial consistency. (iii) {\em Decoder:} It employs transposed convolutional upsampling and convolutional refinement blocks, utilizing skip connections from corresponding encoder layers to reconstruct high-resolution segmentation masks. (iv) {\em Output Layer:} A final $1\times1\times1$ convolution projects decoder outputs into four classes: background, edema, tumor core, and enhancing tumor.

\subsection{TwinSegNet Local Training}

As part of FL architecture, each client independently trains a local TwinSegNet model based on its private MRI dataset. The training workflow is summarized in Algorithm~\ref{alg:twinnet}. To be specific, each client performs local training using its institution-specific data without exposing raw images. As shown in Algorithm~\ref{alg:twinnet}, the training loop consists of multiple epochs, during which data is loaded in batches and preprocessed. For each batch, the model performs a forward pass to predict segmentation outputs, computes a composite loss function based on Dice and Cross-Entropy criteria, and applies backpropagation to update weights. After completing local epochs, the updated model parameters are retained and later transmitted to the central server that conducts aggregation. This process makes sure that learning is tailored to local data distributions while preserving patient privacy. The loss optimization is performed by combining Dice Loss and Cross-Entropy Loss that uses the Adam optimizer, along with a learning rate of $10^{-4}$.

\begin{algorithm}[h]
\caption{TwinSegNet Local Training at Each Client}
\label{alg:twinnet}\footnotesize
\begin{algorithmic}[1]
\REQUIRE MRI modalities $\{T1, T1ce, T2, FLAIR\}$, Ground Truth Segmentation Mask
\STATE Initialize TwinSegNet model parameters $\theta$
\FOR{each local epoch $e=1$ to $E$}
    \FOR{each batch $(x, y)$ from local dataset}
        \STATE Preprocess $x$: intensity normalization, resizing, z-normalization
        \STATE Forward pass through encoder CNN blocks
        \STATE Apply 3D Patch Embedding ViT at bottleneck
        \STATE Forward pass through decoder CNN blocks with skip connections
        \STATE Generate predicted segmentation mask $\hat{y}$
        \STATE Compute composite loss: $L = \text{DiceLoss}(\hat{y}, y)$
        \STATE Backpropagate and update model parameters $\theta$
    \ENDFOR
\ENDFOR
\STATE Return locally updated model $\theta$
\end{algorithmic}
\end{algorithm}

\begin{figure*}
\centering
 \begin{minipage}{.32\textwidth}
  \includegraphics[width=\textwidth,height=\textheight,keepaspectratio]{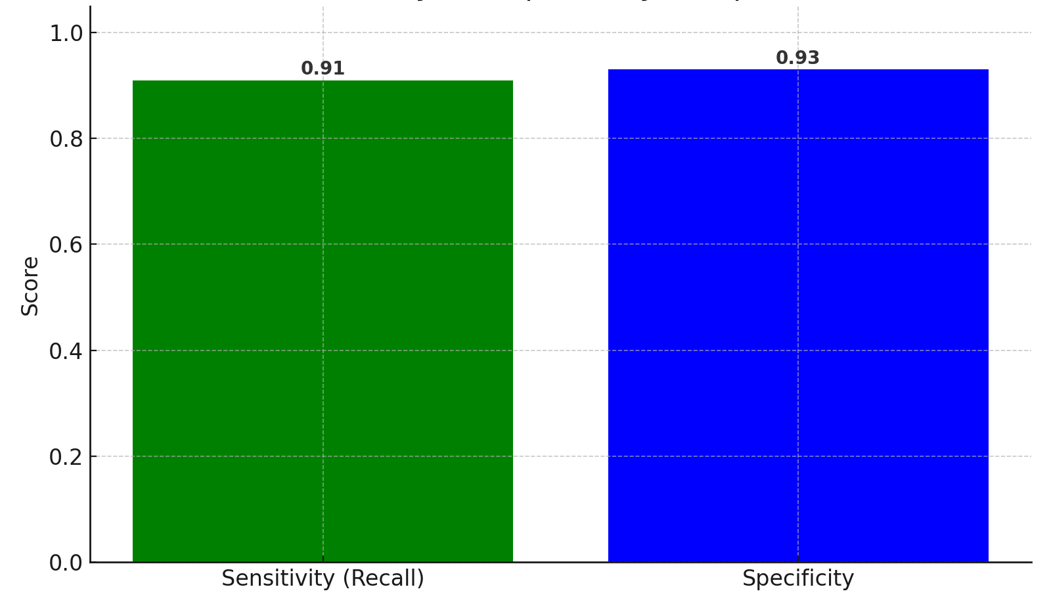}
    \caption{Sensitivity and specificity comparison.}
    \label{fig:sens_spec}
 \end{minipage}
 \begin{minipage}{.32\textwidth}
  \includegraphics[width=\textwidth,height=\textheight,keepaspectratio]{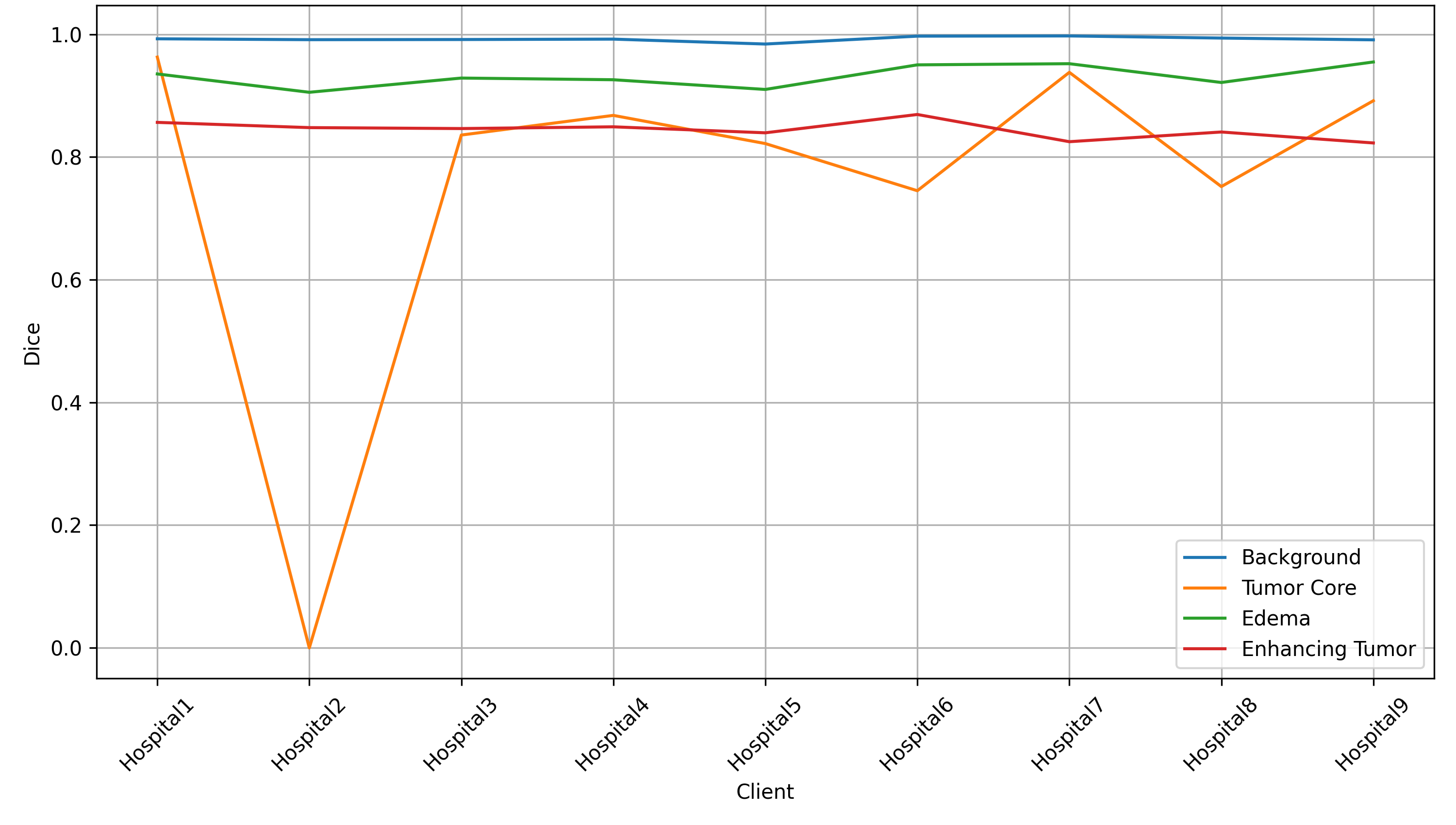}
    \caption{Dice Score per Class Across Nine Hospitals.}
    \label{fig:dice_class}
 \end{minipage}
  \begin{minipage}{.32\textwidth}
  \includegraphics[width=\textwidth,height=\textheight,keepaspectratio]{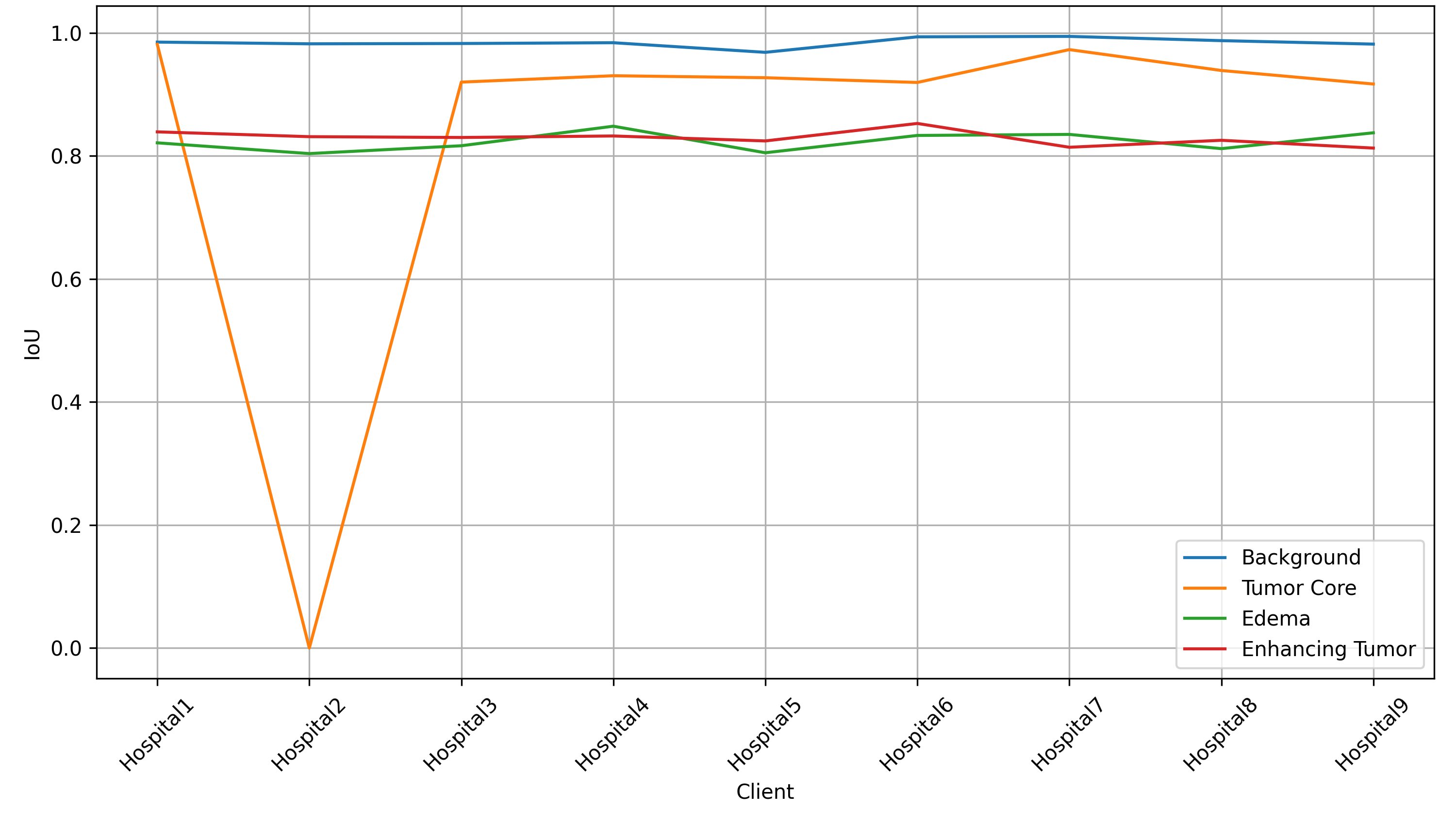}
    \caption{IoU Score per Class Across Nine Hospitals.}
    \label{fig:iou_class}
 \end{minipage}
\end{figure*}

\subsection{Federated Learning Framework}

TwinSegNet is trained under a privacy-preserving FL framework based on the Federated Averaging (FedAvg) algorithm. No raw data is shared across the hospitals; each hospital trains the model locally. The model updates are transmitted to a central server after every local training round for aggregation. The federated training procedure is summarized in Algorithm~\ref{alg:fedavg}.

The global TwinSegNet model is trained collaboratively using the FedAvg algorithm. At the beginning of each communication round, the current global model is shared with all participating clients. Each client trains the received model locally and produces an updated copy of the weights. These local model updates are sent back to the central server, which carries out a weighted average to compute the new global model. This process is repeated across multiple rounds, allowing the global model to progressively integrate knowledge from all clients without accessing any private patient data.

\subsection{Digital Twin Integration}

Each hospital maintains a personalized DT in our implementation by fine-tuning the global model based on its own local data after each federated learning/update round. This lightweight adaptation enables hospital-specific models that reflect their internal patient population without retraining from scratch.

More formally, after the global model $\theta_g$ is aggregated via FedAvg, each client $k$ initializes its DT model $\theta_{dt}^k$ from $\theta_g$ and updates it using one or more epochs of additional training on local data:
$\theta_{dt}^k = \text{FineTune}(\theta_g, D_k)$, where $D_k$ is the private dataset at client $k$. This approach allows each DT to evolve as a locally optimized model tailored to its institutional data distribution.

This DT personalization strategy supports {\em Real-time inference} with improved generalization to the client’s patients, {\em privacy preservation}, as raw data remains local, and {\em Computational efficiency}, requiring only minimal fine-tuning.

\begin{algorithm}[h]
\caption{Federated Averaging (FedAvg)}
\label{alg:fedavg}\footnotesize
\begin{algorithmic}[1]
\STATE Initialize global model $\theta_0$
\FOR{each communication round $r=1$ to $R$}
    \FOR{each client $k$ in parallel}
        \STATE Download global model $\theta_r$ to client $k$
        \STATE Client $k$ trains on local data for $E$ epochs, obtaining updated model $\theta_r^k$
    \ENDFOR
    \STATE Server aggregates updates using dataset-size weights:
    \[
    \theta_{r+1} = \sum_{k=1}^{K} \frac{n_k}{N} \theta_r^k,
    \]
    where $n_k$ is the number of samples at client $k$ and $N = \sum_{k=1}^{K} n_k$.
\ENDFOR
\end{algorithmic}
\end{algorithm}
As shown in Algorithm~\ref{alg:fedavg}, our approach leverages synchronous FL, where all clients train for $E$ epochs per round. In practice, late or missing updates can be handled through partial participation or by weighting delayed updates, ensuring robustness without altering the aggregation process.

\section{Performance Evaluation}
\label{sec:evaluation}

To evaluate TwinSegNet, we conducted experiments across nine federated clients using heterogeneous brain tumor MRI datasets. The global model was trained over 10 federated rounds, with local fine-tuning for personalized DT models. Our FL framework, implemented in PyTorch 1.13, was run on a Lambda server (NVIDIA RTX 3080 GPUs, Intel i9 CPU, 64 GB RAM, Ubuntu 20.04). A simulation-based architecture represented all clients and the server as logical nodes in a single environment, ensuring reproducibility.

Each TwinSegNet model was trained for five local epochs per round with the Adam optimizer ($10^{-4}$ learning rate). MRI volumes were preprocessed to $128 \times 128 \times 128$ resolution with four modalities (T1, T1ce, T2, FLAIR). Performance was evaluated using Dice Similarity Coefficient (DSC), Intersection-over-Union (IoU), sensitivity, and specificity across enhancing tumor (ET), tumor core (TC), and whole tumor (WT) subregions. Qualitative results included visualization overlays comparing predicted and ground-truth masks.

\begin{figure}[htbp]
    \centering
    \includegraphics[width=0.8\linewidth]{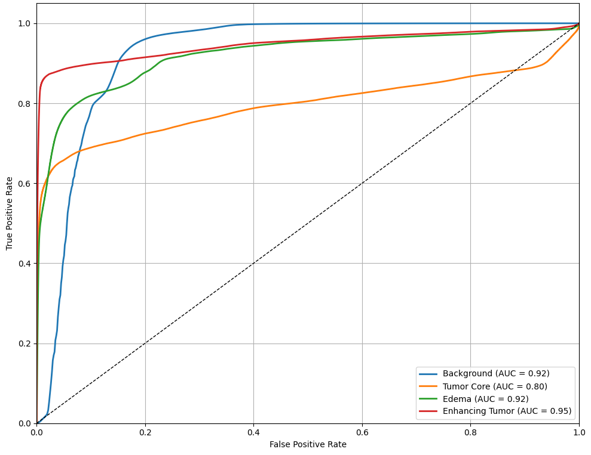}
    \caption{ROC per tumor class showing AUC.}
    \label{fig:roc_curve}
\end{figure}

\begin{figure}[ht]
    \centering
    \includegraphics[width=0.8\linewidth]{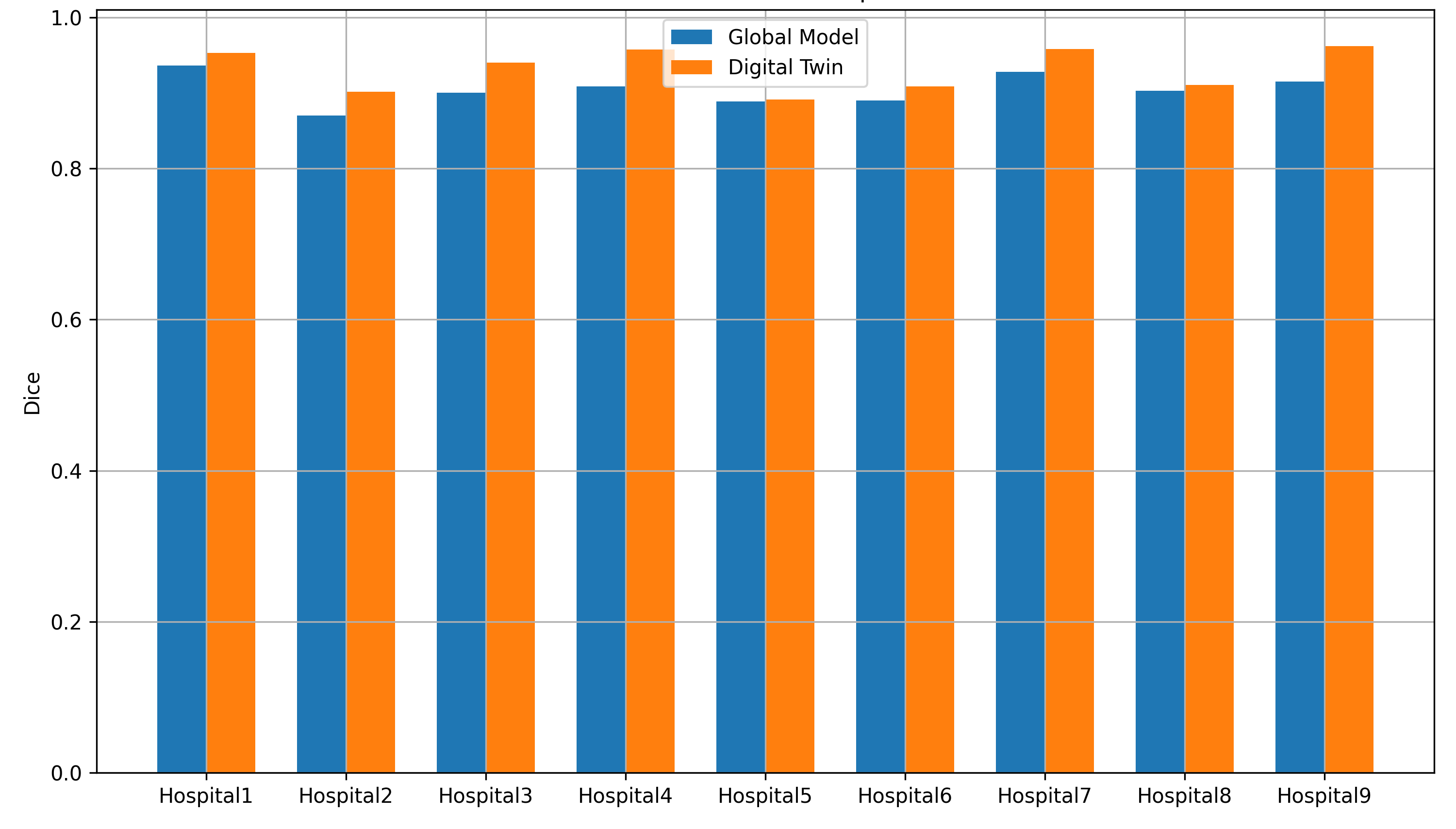}
    \caption{Comparison of Dice Scores: Global vs. DT.}
    \label{fig:dt_dice}
\end{figure}

\begin{figure}[ht]
    \centering
    \includegraphics[width=0.8\linewidth]{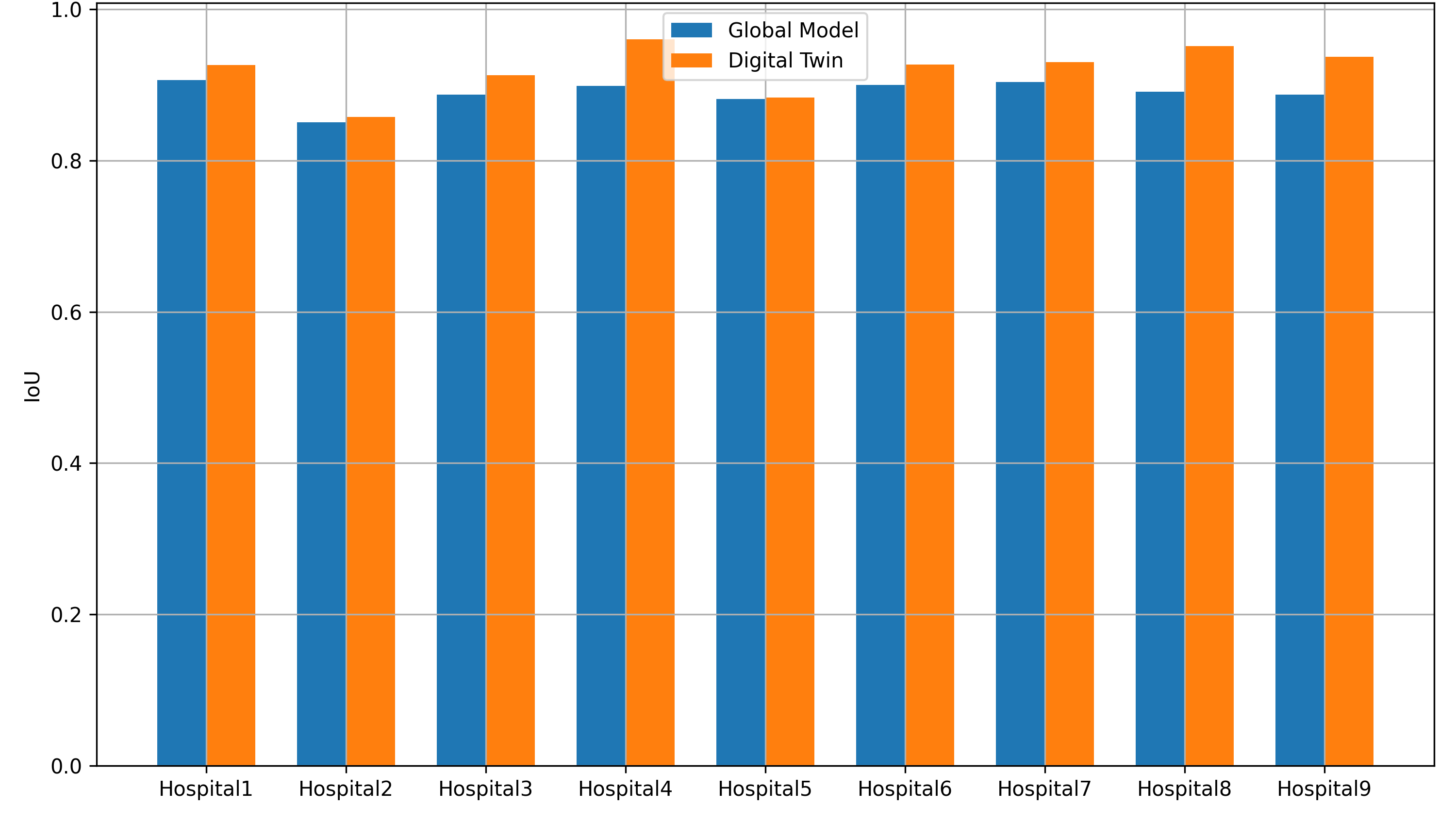}
    \caption{Comparison of IoU Scores: Global vs. DT.}
    \label{fig:dt_iou}
\end{figure}

\begin{table}[ht]
\centering \footnotesize
\caption{Dice and IoU Comparison of Global vs. DT per Client.}
\label{tab:client_summary}
\resizebox{\linewidth}{!}{%
\begin{tabular}{lcccc}
\toprule
\textbf{Client} & \textbf{Global Dice} & \textbf{DT Dice} & \textbf{Global IoU} & \textbf{DT IoU} \\
\midrule
Hospital1 & 0.9368 & 0.9533 & 0.9069 & 0.9246 \\
Hospital2 & 0.8700 & 0.9019 & 0.8509 & 0.8578 \\
Hospital3 & 0.9006 & 0.9401 & 0.8877 & 0.9132 \\
Hospital4 & 0.9087 & 0.9577 & 0.8994 & 0.9606 \\
Hospital5 & 0.8888 & 0.8912 & 0.8841 & 0.8836 \\
Hospital6 & 0.8903 & 0.9089 & 0.8999 & 0.9271 \\
Hospital7 & 0.9279 & 0.9583 & 0.9042 & 0.9303 \\
Hospital8 & 0.9032 & 0.9105 & 0.8910 & 0.9517 \\
Hospital9 & 0.9150 & 0.9622 & 0.8874 & 0.9371 \\
\bottomrule
\end{tabular}
}
\end{table}

\begin{figure}[htbp]
    \centering
    \includegraphics[width=0.9\linewidth]{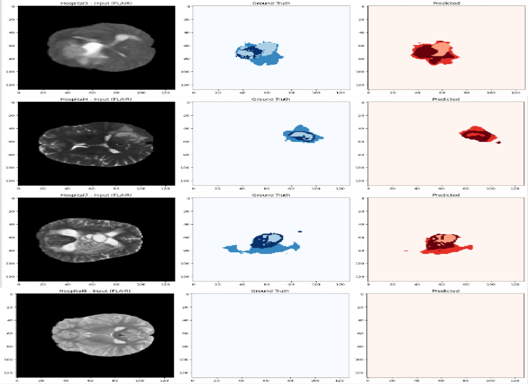}
    \caption{Segmentation results from four clients (Left: Input FLAIR slice, Middle: Ground truth, Right: Predicted mask).}
    \label{fig:overlay}
\end{figure}

\begin{table}[h]
\centering \footnotesize
\caption{Class-wise Dice Coefficient and Mean Dice Comparison.}
\label{tab:classwise-comparission}
\resizebox{0.5\textwidth}{!}{%
\begin{tabular}{lcccc}
\toprule
\textbf{Model} & Dice ET & Dice TC & Dice WT &  Mean Dice \\
\midrule
\cite{chen2021transunet}      & 0.812 & 0.855 & 0.917 & 0.86 \\
\cite{luu2021extending}       & 0.845 & 0.878 & 0.928 & 0.88 \\
\cite{kotowski2021coupling}   & 0.820 & 0.878 & 0.927  & 0.87 \\
\textbf{TwinSegNet (Global)}  & 0.844 & 0.851 & 0.931  & 0.87 \\
\textbf{TwinSegNet (DT)}      & \textbf{0.864} & 0.881 & \textbf{0.940} & \textbf{0.90} \\
\bottomrule
\end{tabular}
}
\end{table}

We report a sensitivity (recall) of 0.91 and a specificity of 0.93, confirming the model’s high ability to recognize true tumor voxels while avoiding false positives, as shown in Fig.~\ref{fig:sens_spec}.
Fig.~\ref{fig:dice_class} and Fig.~\ref{fig:iou_class} show the Dice and IoU scores for each of the four tumor classes across the nine hospitals. The background class consistently achieves high scores ($\textit{Dice} > 0.98$), indicating accurate delineation of non-tumor regions. The Edema class performs robustly across clients, with Dice scores ranging from 0.90 to 0.95. Tumor Core shows more variation. For example, Hospital 2 displays a notable drop due to missing or underrepresented class instances. Enhancing Tumor scores average around 0.85 across all clients, demonstrating stable generalization for this challenging class.

The Receiver Operating Characteristic (ROC) curves in Fig.~\ref{fig:roc_curve} exhibit high Area Under the Curve (AUC) scores for all tumor classes. For instance, 0.92 for Background, 0.80 for Tumor Core, 0.92 for Edema, and 0.95 for Enhancing Tumor, highlighting the model’s strong voxel-level classification capability.

We also assess the DT model personalization by comparing it with the global model. As seen in Fig.~\ref{fig:dt_dice} and Fig.~\ref{fig:dt_iou}, DT models consistently outperform the global model across clients. Hospitals 9 and 3 exhibit up to 4–5\% Dice gains, with average improvements of 1.8\% (Dice) and 1.6\% (IoU), demonstrating that client-specific fine-tuning enhances local performance while preserving privacy.

Table~\ref{tab:client_summary} shows that personalized DT models consistently outperform the global model across clients. Hospitals 4 and 9 exhibit the most significant gains, highlighting the value of personalization in privacy-preserving federated clinical settings.

Fig.~\ref{fig:overlay} compares input FLAIR images, ground-truth masks, and TwinSegNet predictions across four clients. TwinSegNet achieves precise tumor boundary segmentation with strong alignment to ground truth, demonstrating robustness, generalization, and suitability for real-world clinical use.

To benchmark TwinSegNet, we compare class-wise Dice scores and average segmentation accuracy against state-of-the-art centralized models, as shown in Table~\ref{tab:classwise-comparission}. The Global TwinSegNet matches leading methods (mean Dice \textbf{0.875}) while preserving privacy. After local fine-tuning, the DT version achieves the highest mean Dice (\textbf{0.897}), with notable gains in Enhancing Tumor (0.864) and Whole Tumor (0.940). These results confirm that client-specific personalization in a federated setup enhances segmentation accuracy while ensuring data privacy.

\section{Final Remarks}
\label{sec:conclusion}

In this paper, we designed TwinSegNet, a privacy-preserving federated deep learning framework for personalized brain tumor segmentation. Through integrating a hybrid ViT-UNet model with local digital twin personalization, TwinSegNet achieves high segmentation accuracy across multiple institutions without requiring the sharing of raw patient data. Our extensive evaluations on nine heterogeneous hospital datasets confirm the model’s ability to generalize across clients while maintaining high sensitivity and specificity. The integration of local fine-tuning via DT enhances both robustness and clinical relevance. Future work will investigate network-aware optimizations and adaptive communication strategies to further improve the efficiency, scalability, and robustness of DT-enabled FL systems implemented and deployed across diverse healthcare networks. Additional research directions include longitudinal modeling, domain adaptation, and incorporating clinical feedback to enable the continuous refinement and personalization of DT models.

\bibliographystyle{IEEEtran}
\bibliography{references}

\end{document}